\documentclass{article} 
\usepackage{fullpage}

\usepackage{wrapfig,lipsum,booktabs}
\usepackage[utf8]{inputenc} 
\usepackage[T1]{fontenc}    
\usepackage{url}            
\usepackage{booktabs}       
\usepackage{amsfonts}       
\usepackage{nicefrac}       
\usepackage{microtype}      
\usepackage{soul}
\usepackage{graphicx}
\usepackage{amsmath}
\usepackage{outlines}
\usepackage{xurl}

\usepackage[colorlinks=true,
citecolor=blue,
filecolor=black,
linkcolor=blue,
urlcolor=blue]{hyperref}

\usepackage{amsmath,amsfonts,bm}

\def\eqref#1{equation~\ref{#1}}

\def\1{\bm{1}}

\DeclareMathAlphabet{\mathsfit}{\encodingdefault}{\sfdefault}{m}{sl}
\SetMathAlphabet{\mathsfit}{bold}{\encodingdefault}{\sfdefault}{bx}{n}

\usepackage{multirow}
\usepackage{paralist}

\usepackage{multicol}
\usepackage{diagbox}

\usepackage[ruled,noend]{algorithm2e}

\SetCommentSty{mycommfont}

\usepackage{here}

\usepackage{amsmath,amssymb,amsfonts,amsbsy,amsfonts,latexsym}
\usepackage{makecell}
\usepackage{xcolor}
\usepackage{colortbl}

\usepackage{tabularx,colortbl,xcolor}
\usepackage[normalem]{ulem}
\useunder{\uline}{\ul}{}

\usepackage{enumitem}

\usepackage{xparse}

\SetKwInput{KwInput}{Input}
\SetKwInput{KwRequire}{Require}

\usepackage{longtable}

\NewDocumentCommand{\var}{O{s} m O{}}{%
  \ensuremath{#1_{#2}^{#3}}
}
\usepackage{siunitx}

\newcommand{\commentout}[1]{}

\definecolor{light-gray}{gray}{0.80}

\newcommand\appref{Appendix~\ref}

\newcommand\fref{Figure~\ref}
\newcommand\tref{Table~\ref}
\newcommand\sref{Section~\ref}

\usepackage{amsthm}

\usepackage{amsthm}

\newtheorem{mytheorem}{Theorem}
\newtheorem{assumption}[mytheorem]{Assumption}
\newtheorem{lemma}[mytheorem]{Lemma}
\newtheorem{remk}[mytheorem]{Remark}

\usepackage{subfig}

\newcommand{\ppl}{PPL\xspace}

\newcommand{\opt}{OPT\xspace}
\newcommand{\bloom}{BLOOM\xspace}
\newcommand{\llama}{LLaMA\xspace}

\usepackage{pifont}
\newcommand{\cmark}{\ding{51}}%
\newcommand{\xmark}{\ding{55}}%

\usepackage{arydshln}
\usepackage{xcolor}
\usepackage{listings}
\definecolor{myblue}{rgb}{0,0.2,0.8}
\hypersetup{ %
pdftitle={},
pdfauthor={},
pdfsubject={},
pdfkeywords={},
pdfborder=0 0 0,
pdfpagemode=UseNone,
colorlinks=true,
linkcolor=myblue,
citecolor=myblue,
filecolor=myblue,
urlcolor=myblue,
pdfview=FitH}
\makeatletter
\def\adl@drawiv#1#2#3{%
        \hskip.5\tabcolsep
        \xleaders#3{#2.5\@tempdimb #1{1}#2.5\@tempdimb}%
                #2\z@ plus1fil minus1fil\relax
        \hskip.5\tabcolsep}
\newcommand{\cdashlinelr}[1]{%
  \noalign{\vskip\aboverulesep
           \global\let\@dashdrawstore\adl@draw
           \global\let\adl@draw\adl@drawiv}
  \cdashline{#1}
  \noalign{\global\let\adl@draw\@dashdrawstore
           \vskip\belowrulesep}}
\makeatother

\definecolor{upforestgreen}{rgb}{0.6, 0.8, 0.2}

\usepackage{chngcntr}
\usepackage{adjustbox}
\usepackage{wrapfig}
\usepackage{arydshln}

\usepackage{tcolorbox}

\begin{document}

\title{ZeroQuant-FP: A Leap Forward in LLMs Post-Training W4A8 Quantization Using Floating-Point Formats}

\author{
 Xiaoxia Wu\thanks{Equal Contribution. Code will be released  as a part of \url{https://github.com/microsoft/DeepSpeed}} ,    Zhewei Yao$^*$,    Yuxiong He
\\  Microsoft \\ 
{\tt \small\{zheweiyao,  xiaoxiawu, yuxhe\}@microsoft.com}
}

\date{}
\maketitle

\begin{abstract}
In the complex domain of large language models (LLMs), striking a balance between computational efficiency and maintaining model quality is a formidable challenge. 
Navigating the inherent limitations of uniform quantization, particularly when dealing with outliers, and motivated by the launch of NVIDIA's H100 hardware, this study delves into the viability of floating-point (FP) quantization, particularly focusing on FP8 and FP4, as a potential solution. 
Our comprehensive investigation reveals that for LLMs, FP8 activation consistently outshines its integer (INT8) equivalent, with the performance edge becoming more noticeable in models possessing parameters beyond one billion. 
For weight quantization, our findings indicate that FP4 exhibits comparable, if not superior, performance to INT4, simplifying deployment on FP-supported hardware like H100.  To mitigate the overhead from precision alignment caused by the disparity between weights and activations, we propose two scaling constraints for weight quantization that negligibly impact the performance compared to the standard W4A8 model. We additionally enhance our quantization methods by integrating the Low Rank Compensation (LoRC) strategy, yielding improvements especially in smaller models. The results of our investigation emphasize the immense potential of FP quantization for LLMs, paving the way for high-efficiency deployment in resource-limited settings.
\end{abstract}
\section{Introduction}


As Natural Language Processing (NLP) evolves, Large Language Models (LLMs) like Codex~\cite{copilot} and ChatGPT~\cite{chatgpt} have become essential, transforming our interaction with technology and daily communication. However, their complexity and computational intensity present deployment challenges ~\cite{pope2022efficiently,gholami2020ai,smith2022using}, particularly in resource-limited settings. One solution is quantization, which represents data in lower-precision formats such as 8-bit integers or floating-point numbers, reducing memory needs and potentially enhancing inference latency through better GEMM computation throughput on compatible GPUs. Post-Training Quantization (PTQ), which directly reduces the precision of a fully trained model's parameters, is often preferred for LLMs due to its simplicity and lower computational overhead.\footnote{Note that we do not discuss Quantize-Aware Training (QAT) for LLMs in this paper as QAT require the computation graph for back-propogation~\cite{kim2016bitwise,banner2018scalable,mellempudi2017ternary,wu2023understanding,wu2022extreme,dettmers2023qlora,liu2023llm}.} Recent studies indicate that PTQ on 8-bit integer (INT8) weight-only quantization does not compromise the quality of LLMs~\cite{yao2022zeroquant,dettmers2022llm,xiao2022smoothquant,wei2023outlier}, and only a minor accuracy drop is observed with INT4 weight quantization when advanced algorithm such as GPTQ  applied~\cite{frantar2022gptq,yao2023comprehensive,kim2023squeezellm,lin2023awq}.

The exploration of activation quantization, in addition to weight-only quantization, has also gained interest. This approach expedites inference times by taking advantage of unified precision leading to more efficient execution on hardware. The primary challenge in implementing activation quantization lies in the trade-off between efficiency and performance. As evidenced in studies such as ZeroQuants~\cite{yao2022zeroquant,yao2023comprehensive}, SmoothQuant~\cite{xiao2022smoothquant} and others, reducing the precision of activation from FP16 to INT8 inevitably results in a decrease in model quality. This degradation is partially due to the presence of extreme values or outliers in the activation of LLMs \cite{dettmers2022case,xiao2022smoothquant,lin2023awq,kim2023squeezellm}, which is partly attributed to the pretraining  effect \cite{wu2023understanding}. In the presence of outliers, uniform quantization like INT8 or INT4, fail to accurately represent the main body of the data as they become skewed towards the outlier. This issue stems from the inherent assumption in these techniques of a uniform data distribution~\cite{wu2020integer}, an assumption that might not correspond to the actual data points distribution.


Considering the drawbacks of integer quantization delineated previously, floating-point (FP) methods like FP8 or FP4, employing ExMy notation, arise as more potent alternatives~\cite{micikevicius2022fp8,cambier2020shifted,kuzmin2022fp8,van2023fp8,zhang2023integer}.  Unlike the fixed range of integer types, floating-point methods allow for adjusting the decimal point position, enabling dynamic scaling across activation maps and preserving important features. While there is debate about the quality of models between integer and floating-point quantization~\cite{van2023fp8}, recent research on PTQ LLMs using FP8/FP4 in \cite{zhang2023integer}  reveals FP8 to be substantially better than INT8 activation quantization. In terms of hardware support and performance, while INT8 computations are broadly supported by most modern CPUs and GPUs \cite{fastertransformer,wu2023understanding}, lower-bit floating-point operations are also  increasingly recognized in the industry. An example of this is the newly release of NVIDIA's H100 GPU, specifically engineered for FP8 computations~\cite{micikevicius2022fp8}. Hence, despite the potentially higher computation cost of FP8 compared to INT8 and in light of hardware support, the improved model quality could make this trade-off worthwhile and merits further exploration.

While a few studies such as the one by \cite{zhang2023integer} have ventured into the realm of post-training FP quantization in LLMs, they have unveiled considerable drawbacks in terms of model quality. Specifically, when implementing GPTQ \cite{frantar2022gptq} for FP8 quantization on both weights and activation for models such as \llama-7B or \llama-30b \cite{touvron2023llama}, there is an observed perplexity degradation surpassing 1.0 on Wiki-text2 dataset~\cite{merity2016pointer}. This level of model degradation presents significant practicality issues, hindering the optimal utilization of these models. In response to these findings, our paper undertakes an in-depth exploration into FP quantization. We primarily focus on the variance in activation values—an integral element that could potentially be the key to enhancing the performance of these quantization techniques. Our main contributions include:

\begin{itemize}
    \item Demonstrating minimal degradation with FP8 activation and weight quantization: Particularly in larger models, FP8 activation and weight quantization result in negligible degradation, performing comparably to the original FP16 models. 
    \item Identifying potential in FP8 activation and FP4 weights, and the impact of Low Rank Compensation (LoRC): We highlight the potential in FP8 activation and FP4 weights. The LoRC method, proposed in \cite{yao2023comprehensive}, significantly reduces quantization errors in the W4A8 scheme for FP quantization, especially in smaller models, thereby enhancing performance.
    
    \item Illustrating the maintenance of quality in the W4A8 floating-point model even when constraints are imposed on the scaling factors: For true efficiency in the W4A8 model, a conversion from FP4 to FP8 for weight is crucial. 
  To alleviate this converting overhead, we here suggest two possible scaling constraints for weight quantization: (1) restricting all scaling factors to be a power of 2 and (2) requiring the scaling factors in one compute group (e.g., several rows of the weight matrix~\cite{yao2022zeroquant} to be transferable by simple bit-shifting).
    Our analysis indicates that these two restrictions negligibly affect the model's performance in comparison to the conventional W4A8 configuration. 
\end{itemize}


\section{Background}
\label{subsec:quantization_challenges}






The impact of 8-bit activation quantization, especially potential accuracy loss, is comprehensively outlined in  ZeroQuant-V2~\cite{yao2023comprehensive}. They present a direct comparison between the W16A16 and W16A8 (INT8) quantization schemes across a variety of models. To provide an easier understanding, we quoted their results in \tref{table:compare-8-16A} for both OPT\cite{zhang2022opt} and BLOOM \cite{scao2022bloom} models, which indicates that the quality of models, especially the OPT family, is significantly influenced by the activation quantization.
 \begin{table}[H]
\caption{Comparison of FP16 and INT8 activation quantization. We report the average \ppl  (the lower the better) over Wikitext-2 (WIKI)~\cite{merity2016pointer}, PTB~\cite{marcinkiewicz1994building}, and C4~\cite{colin2019t5}, for both \opt and \bloom (BLM) models.  }\centering
\label{table:compare-8-16A}
\begin{adjustbox}{width=0.999\linewidth}
\centering
\begin{tabular}{lcccccccccccccc }
\toprule
Precision     &  OPT-6.7b	& OPT-13b &OPT-30b	& OPT-66b   & BLM-1.7b  & BLM-3b & BLM-7.1b & BLM-176b\\
\midrule
W16-A16     & 11.90 & 11.22   & 10.70   & 10.33   & 20.43 & 17.58 & 14.96 & 10.90 \\\midrule
W16-A8 & 12.62 & 15.36   & 23.57   & 561.35  & 20.52 & 17.65 & 15.14 & 11.62\\
\bottomrule
\end{tabular}
\end{adjustbox}
\end{table}
\begin{figure}[H]
\centering
\includegraphics[width=1.0\linewidth]{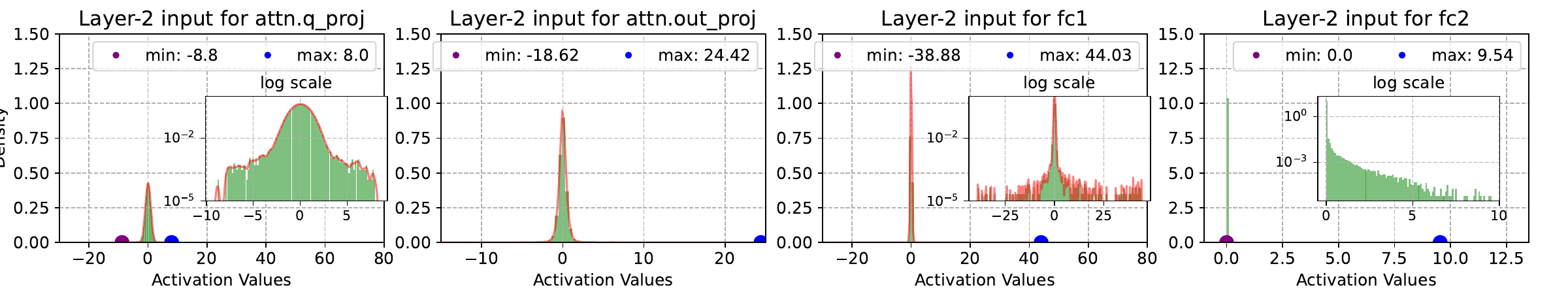}
\includegraphics[width=1.0\linewidth]{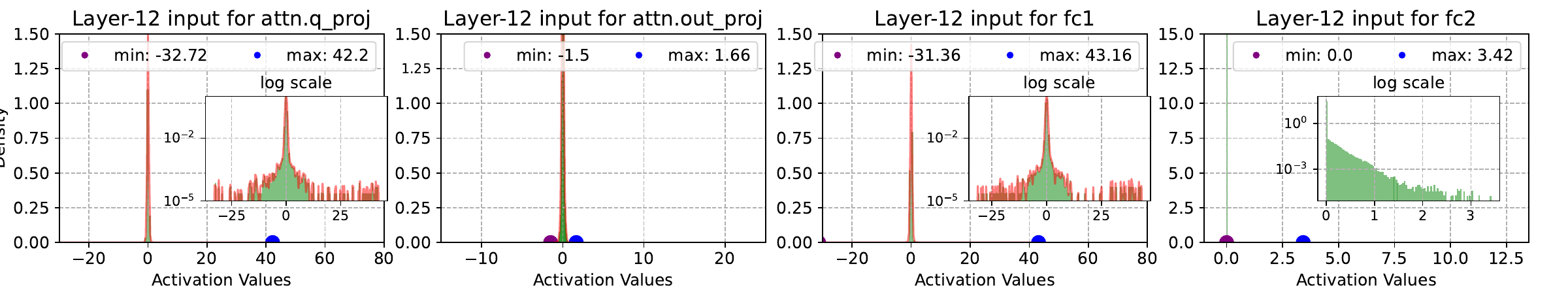}
\includegraphics[width=1.0\linewidth]{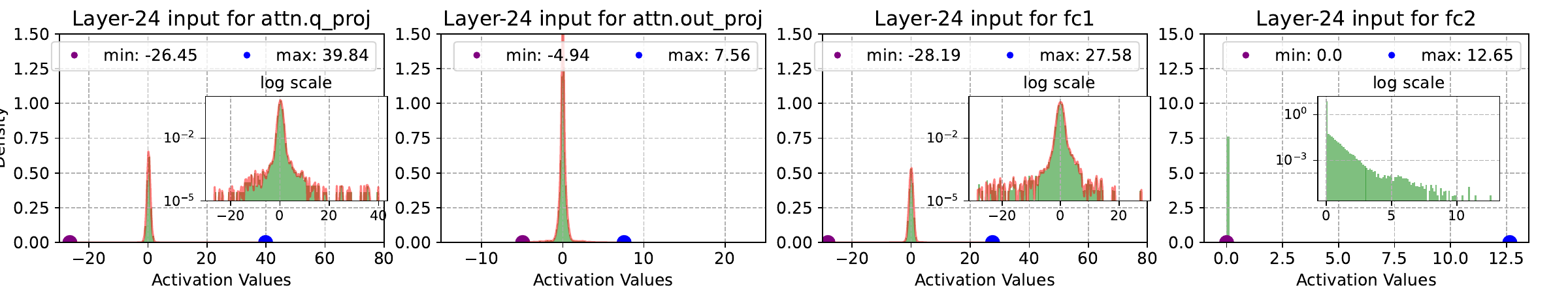}
\vspace{-0.5cm}
\caption{Distribution of Activation values. The top, middle and bottom rows represents the distributions at the 2nd, 12th and final layer of the pretrained OPT-1.3b model. From the left to right columns, they are respectively for the linear modules attn.q\_proj (same as attn.k\_proj and attn.v\_proj), attn.out\_proj, fc1, and fc2. The histogram's x-axis ranges from the smallest to largest activation values, while the y-axis denotes their frequency in the dataset. See legend for their minimum and maximum values. Density functions illustrate the probability of different activation values. For more details, please see \sref{sec:activation}.}
\label{figure:activation}
\end{figure}
\textbf{Distribution of Activations.}\label{sec:activation} We sought to understand the cause of the aforementioned degradation from FP16 activation and INT8,  prompting us to scrutinize the distribution of activation values illustrated in \fref{figure:activation}. We selected a random sentence from the C4 dataset and processed it through a pre-trained OPT-1.3B model. The statistical activation inputs for the 2nd, middle, and final layer were subsequently chosen for a detailed examination. The four histograms correspondingly represent the activations for the Multi-head Attention (MHA) and Multi-Layer Perceptron (MLP) components:\footnote{The hidden dimension for the model is 2048 for `attn.q\_proj', `attn.out\_proj' and `fc1', and 8196 for `fc2'. We pick 20 tokens (position 8 to 28) and vectorize this $20\times 2048$ or $20\times 8196$ matrices to plot their distributions. The plots used bin=100.}
\begin{itemize}
    \item attn.q\_proj, the input for the  query, key, or value in the MHA mechanism,
 \item  attn.out\_proj, the input for the MHA's projection matrices,
 \item fc1, the initial input for the fully-connected  (fc1) projection in MLP,
\item fc2, the subsequent input for the fully-connected (fc2) projection in the MLP.
\end{itemize}

The activation distribution outlined in \fref{figure:activation} reveals some compelling patterns. The input to the  attn.q\_proj module in the 2nd layer (depicted in the 1st column of \fref{figure:activation} in the top row) appears to conform closely to a normal distribution, a result of the layer-normalization process. Yet, moving forward to the subsequent modules within the 2nd layer, namely attn.out\_proj, fc1, and fc2, we notice a skewness in the distributions, with noticeable outlier values. Two distinct observations arise: \textbf{(1)} Regarding the activation distribution for attn.q\_proj and fc1, even though they have undergone layer-normalization, the skewness still presents itself and becomes more conspicuous as we delve deeper into the layers (see the first and third column in the  middle and bottom plots in \fref{figure:activation}). \textbf{(2)} The skewness reaches its peak in the fc2 module. In this particular module, a large portion of the values cluster around zero, with only a handful surpassing this range. This phenomenon is due to the inputs being processed by the "ReLU" (Rectified Linear Unit) operator. This operator, purposefully, voids any negative input values, resulting in a skewed distribution focused around zero. Only positive activation values persist unmodified, giving rise to the outliers observed. This extreme skewness is most noticeable at the final layer (the bottom row in \fref{figure:activation}).

These observations offer a deeper understanding of how activation quantization impacts various modules, even within the same layer. Consequently, this signifies that we must exercise caution when selecting quantization methods. Quantization techniques that employ integers, such as INT8 or INT4, and rely on uniform quantization, may not be ideally suited to manage distributions that are skewed. This is due to the inherent assumption of uniform distribution within these methods, which may not align with the actual distribution of data points.

 \textbf{The Uniform Quantization of INT8.}  The integer quantization such as in INT8 or INT4 states
\begin{equation}
\small
\label{eq:quantization_formula}
Q(x) = \text{INT}\big({(x-Z)}/{S}\big)-Z,
\end{equation}
where $Q$ is the quantization function, $x$ is a floating point input vector/tensor, $S$ is a real valued scaling factor, and $Z$ is an integer zero point. 
Based on different settings, the quantization method can be viewed as  symmetric ($Z=0$) or asymmetric  ($Z\ne0$)  quantization. In scenarios where outliers exist, uniform quantization techniques like INT8 and INT4, regardless of their symmetric or asymmetric variants, frequently fail to accurately approximate the values of clustered data. Consequently, this makes the quantization error larger for those clustered values, as these methods attempt to adjust their fit to accommodate the outlier. Essentially, these techniques become skewed towards the outlier, leading to a reduced accuracy in representing the main body of the data.
\begin{figure}[H]
\centering
\includegraphics[width=1.0\linewidth]{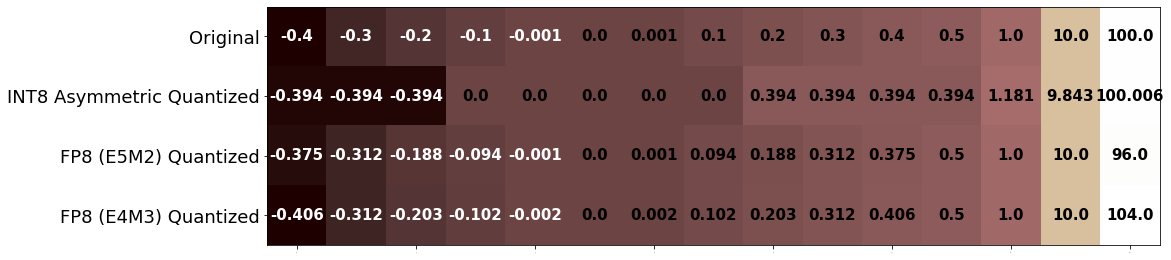}
\vspace{-0.8cm}
\caption{A Contrast between INT8 and FP8 Quantization Methods. The top row displays the original vector in its full-precision form. The subsequent row showcases the vector after quantization through the INT8 Asymmetric approach. The final two rows present values quantized by the FP8 method, utilizing E5M2 and E4M3 formats respectively.}
\label{figure:fp-int}
\end{figure}
\noindent Given the limitations of integer quantization, floating-point methods such as FP8 or FP4, utilizing ExMy notation, emerge as superior alternatives. In these methods, the `x' and `y' values represent the bits allocated for the exponent and mantissa, respectively, totaling to 7 in FP8 or 3 in FP4. The flexibility of FP8 lies in its ability to adjust the decimal point position, unlike integer types with a fixed range.

To demonstrate the disparity between INT8 and FP8, we present \fref{figure:fp-int} where a hypothetical 15-element vector with an outlier value of 100 undergoes quantization using INT8 Asymmetric and FP8 (with both E5M2 and E4M3 configurations). \fref{figure:fp-int} illustrates that while INT8 approximates the outlier effectively, it struggles to accurately represent smaller numbers. Conversely, FP8 (whether with E5M2 or E4M3) provides greater precision in approximating the clustered data.\footnote{Please note that the FP8 format used in this paper is based on the Qtorch Python package, which can be installed via `pip install qtorch'. It differs slightly from Nvidia's FP8 in H100, which requires one mantissa bit-pattern for NaN values.}

Considering the advantages of ExMy, which allows for dynamic scaling across activation maps, quantization error is reduced and essential features are preserved. In this paper, we investigate the performance of FP8 or FP4 techniques for handling the variability in activation or weight values. This could potentially lead to an enhancement in the model's performance on post-training quantization.

\section{Methodology}
\label{sec:methodology}




Several lightweight optimization-based methods, where the weight of the model is updated during quantization, have been proposed in the literature \cite{yao2022zeroquant,yao2023comprehensive,dettmers2022case,xiao2022smoothquant,lin2023awq,kim2023squeezellm}. Among these, we chose to align our approach with the principles outlined in the GPTQ~\cite{frantar2022gptq,frantar2022optimal}, which can be dated back to \cite{lecun1990optimal,hassibi1993second}. While this strategy offers a robust starting point, it is imperative to keep in mind the dynamic and ever-evolving nature of the field of artificial intelligence. There may be more efficient methodologies on the horizon, waiting to be discovered and implemented.  

In light of ZeroQuant-V2~\cite{yao2023comprehensive}, we applied fine-grained quantization (FGQ) for weight and token-wise quantization for activation. In addition, we will also investigate the add-on feature LoRC (Low Rank Compensation) proposed in~\cite{yao2023comprehensive}, which aims to reduce quantization errors in weights by employing low-rank matrix factorization. LoRC involves two main steps: first, it performs Singular Value Decomposition (SVD) on the error matrix, which is the difference between the original weight and the quantized weight. The error matrix is thus decomposed into two unitary matrices and a diagonal matrix. Second, the method formulates a new error approximation using two low-rank matrices that are derived from the matrices in the first step. This approximation is then added to the quantized weight to yield a more accurate estimate of the original weight, thereby reducing quantization errors. 

Based on GPTQ (without or with LoRC), we perform comprehensive comparisons between the use of FP8 or INT8 activation quantization, coupled with adjusting the weight quantization to FP8 and FP4. Particularly we explore the potential of FP4 weight and FP8 activation  quantization.



\textbf{Casting the FP4 to FP8.} \label{sec:fp4tofo8}  Lastly, a unique challenge arises due to the use of different precision levels for weights (W) and activations (A). The actual software implementation of W4A8 in H100 NVIDIA hardware is that one needs to cast  W's  FP4 to match the FP8 precision used in A. The direct method of dequantization followed by quantization again could potentially have a detrimental effect on inference efficiency, hence it is not a viable solution. To address this, we propose the bit-shifting method. 
This means that instead of allowing $S$ defined in \eqref{eq:quantization_formula} to be any real valued scaling factor, we constrain $S$ to be a power of $2$, i.e., $S=2^n, n\in N$ ($S$ could still represent fractions when n is negative and whole numbers when n is not negative). There are two methods we will implement:
\begin{itemize}
    \item[\textbf{(M1)}] Map to the nearest values represented by the power of 2, i.e., letting the new scale $\hat{S} = 2^{\lceil \log_2(S)\rceil}$;
    \item[\textbf{(M2)}] Collect scales to form a vector  $\mathbf{S} = [S_1, S_2, \ldots, S_n]$. Take the maximum value in this group (usually, the set consists a (multiple) row(s) of a matrix~\cite{yao2022zeroquant}), denoted as $S_{\max}$, adjust these elements   $S_{\max}/S_i$  to be represented by the power of 2,  and then define $\hat{S}_i = S_{\max}/ 2^{\lceil \log_2(S_{\max}/S_i)\rceil}$. This provides a far superior approximation compared to (M1).\footnote {To ensure the casting of F4-E2M1 for each weight matrix to FP8, we apply format E5M2 once a matrix is quantized.}
\end{itemize}
 We reiterate that this restriction using the power of 2, either using (M1) or (M2), simplifies computations, especially in digital systems operating  based on binary logic. This is a crucial element of our approach to optimizing computational efficiency and maintaining the performance of our model.

\section{Main Results}
\label{sec:results}

\begin{table}[t] 
\caption{
The evaluation outcomes for \llama (top) and  \opt (bottom ) using different Integer (INT) and Floating-point (FP) quantization methods applied to weight and activation. The performance is measured in terms of perplexity (lower scores are better) and spans across three datasets: WikiText-2 (WIKI), PTB, and C4. For each model, the results initially highlight the average performance across the datasets, followed by a detailed breakdown of outcomes per dataset.
}\centering
\label{tab:main-fp}
\begin{adjustbox}{width=0.999\linewidth}
\centering
\begin{tabular}{ll|cc|cc|cc|cccccc }
\toprule
\multirow{2}{*}{Q-type}       &  Weight-   & \multicolumn{2}{c|}{\llama-3b}  & \multicolumn{2}{c|}{\llama-7b}  & \multicolumn{2}{c|}{\llama-13b}  & \multicolumn{2}{c}{\llama-30b} \\
    &  -Activation & Mean & WIKI/PTB/C4  & Mean & WIKI/PTB/C4  &  Mean & WIKI/PTB/C4 &  Mean & WIKI/PTB/C4\\
\midrule
W16A16& N/A         & 11.93 & 7.35/19.1/9.34  & 13.37 & 5.68/27.35/7.78 & 10.31 & 5.09/19.22/6.61 & 5.79 & 4.10/7.30/5.98   \\ \midrule
\multirow{3}{*}{W8A8}          & INT -- INT & 12.00 & 7.41/19.16/9.41 & 13.58 & 5.72/27.89/7.13 & 10.63 & 5.16/20.07/6.67 & 5.90 & 4.21/7.42/6.06 \\
                               & INT -- FP  & 11.96 & 7.37/19.16/9.35 & 13.45 & 5.69/27.57/7.09 & 10.38 & 5.11/19.42/6.62 & 5.80 & 4.11/7.31/5.99 \\
                               & FP -- FP   & 11.99 & 7.37/19.23/9.37 & 13.46 & 5.70/27.58/7.10 & 10.38 & 5.11/19.41/6.62 & 5.81 & 4.12/7.31/5.99 \\ \midrule
\multirow{3}{*}{W4A8}          & INT -- INT & 12.55 & 7.67/20.23/9.74 & 16.23 & 6.44/34.45/7.79 & 11.48 & 5.32/22.35/6.78 & 6.02 & 4.36/7.54/6.16 \\
                               & INT -- FP  & 12.39 & 7.62/19.87/9.68 & 16.09 & 6.75/33.80/7.72 & 11.31 & 5.28/21.91/6.73 & 5.94 & 4.27/7.45/6.11 \\
                               & FP -- FP   & 12.45 & 7.62/20.05/9.67 & 15.14 & 6.32/31.61/7.51 & 11.08 & 5.26/21.27/6.73 & 5.92 & 4.26/7.42/6.09 \\
                               \midrule
\multirow{2}{*}{W4A8} & INT -- INT & 12.52 & 7.65/20.18/9.72 & 14.14 & 5.88/29.26/7.27 & 10.81 & 5.28/20.38/6.76 & 6.00 & 4.34/7.51/6.14 \\
\multirow{2}{*}{+LoRC}                                & INT -- FP  & 12.38 & 7.58/19.89/9.65 & 14.01 & 5.84/28.95/7.24 & 10.56 & 5.22/19.75/6.71 & 5.90 & 4.24/7.39/6.07 \\
                               & FP -- FP   & 12.42 & 7.61/19.98/9.66 & 13.95 & 5.87/28.75/7.24 & 10.80 & 5.24/20.46/6.72 & 5.91 & 4.26/7.40/6.07\\
\bottomrule
\end{tabular}
\end{adjustbox}

\begin{adjustbox}{width=0.999\linewidth}
\centering

\begin{tabular}{ll|cc|cc|cc|cccccc }
\toprule
\multirow{2}{*}{Q-type}         &  Weight  --   & \multicolumn{2}{c|}{\opt-3b}  & \multicolumn{2}{c|}{\opt-7b}  & \multicolumn{2}{c|}{\opt-13b}  & \multicolumn{2}{c}{\opt-30b} \\
    &  -- Activation & Mean & WIKI/PTB/C4  & Mean & WIKI/PTB/C4  &  Mean & WIKI/PTB/C4 &  Mean & WIKI/PTB/C4\\
\midrule
W16A16   & N/A& 15.44 &14.62/16.97/14.72 & 11.90 & 10.86/13.09/11.74 &11.22 &10.13/12.34/11.20  & 10.70 &9.56/11.84/10.69
\\
\midrule
\multirow{3}{*}{W8A8}          & INT -- INT & 15.94 & 14.98/17.49/15.36 & 12.66 & 11.20/14.29/12.48 & 15.94 & 12.13/19.82/15.86 & 25.76 & 14.63/32.90/29.74 &  \\
                               & INT -- FP  & 15.85 & 14.93/17.56/15.05 & 11.99 & 10.92/13.24/11.80 & 11.27 & 10.16/12.42/11.23 & 10.69 & 9.51/11.87/10.71  &  \\
                               & FP -- FP   & 15.86 & 14.97/17.55/15.05 & 11.99 & 10.91/13.24/11.81 & 11.27 & 10.16/12.42/11.23 & 10.69 & 9.51/11.87/10.71  &  \\\midrule
\multirow{3}{*}{W4A8}          & INT -- INT & 16.41 & 15.39/18.22/15.62 & 13.18 & 11.61/15.00/12.92 & 16.70 & 12.32/21.21/16.56 & 24.42 & 14.80/30.38/28.09 &  \\
                               & INT -- FP  & 16.40 & 15.46/18.23/15.51 & 12.20 & 11.13/13.49/11.99 & 11.34 & 10.20/12.53/11.30 & 10.73 & 9.54/11.91/10.75  &  \\
                               & FP -- FP   & 16.29 & 15.32/18.19/15.35 & 12.09 & 10.89/13.44/11.95 & 11.34 & 10.16/12.55/11.30 & 10.72 & 9.52/11.90/10.75  &  \\\midrule
\multirow{2}{*}{W4A8} & INT -- INT & 16.38 & 15.50/18.05/15.59 & 12.75 & 11.37/14.33/12.53 & 15.89 & 12.06/19.76/15.85 & 27.20 & 15.94/34.50/31.16 &  \\
\multirow{2}{*}{+LoRC}  & INT -- FP  & 16.23 & 15.40/17.97/15.32 & 12.13 & 11.07/13.43/11.90 & 11.34 & 10.23/12.49/11.29 & 10.71 & 9.48/11.91/10.74  &  \\
                               & FP -- FP   & 16.23 & 15.50/17.92/15.28 & 12.09 & 10.96/13.40/11.90 & 11.33 & 10.15/12.55/11.29 & 10.71 & 9.48/11.90/10.75  &                       
                      \\
\bottomrule
\end{tabular}
\end{adjustbox}
\end{table}

In this section, we perform experiments to understand the differences of Integer (INT) and Floating-point (FP) quantization using the GPTQ methods~\cite{frantar2022gptq} with or without the add-on feature LoRC~\cite{yao2023comprehensive}. As described in \sref{subsec:quantization_challenges}, floating-point quantization could potentially maintain more precise information, which might improve the model's performance. To see if this is true, we include two model-type families: \llama~\cite{touvron2023llama} and \opt~\cite{zhang2022opt}, with sizes ranging from  1 billion to 30 billion parameters.  The evaluation spans across three datasets: Wikitext-2 (WIKI)~\cite{merity2016pointer}, PTB~\cite{marcinkiewicz1994building}, and C4~\cite{colin2019t5}.   For more experiment details, please see \appref{sec:experiment-details}.

The primary results in Table \ref{tab:main-fp} reveal the impact of various quantization types which are  applied to weight and activation specified in the 2nd column; for instance, W4A8 precision, INT -- FP means INT4 is used for weight and FP8 for activation. Our results provide an average performance over three datasets, offering a broad understanding of the quantization method's efficiency. However, we delve further, understanding that these methods' performance can differ with the characteristics of datasets, thus presenting a detailed performance breakdown for each dataset.  We find that FP8 and FP4, the configurations E4M3 and E2M1 respectively outperform E5M2 and E3M0, hence, they were used in our experiments. Further insights and explanations regarding these configurations' impact on performance are to be addressed in \appref{sec:experiment-details}.

\textbf{FP8 Activation is much better than INT8.} The high-level summary of the results in Table \ref{tab:main-fp} indicates that for both \llama and \opt model families, FP8 activation generally outperforms INT8 activation. This observation corroborates the motivation discussed in Section \ref{subsec:quantization_challenges}, emphasizing FP8's superior capacity to capture more nuanced information, a vital aspect for generative tasks in large-scale LLMs.

Interestingly, the advantage of FP8 over INT8 becomes more pronounced for larger models with parameters greater than 6.7 billion, such as \llama-7b/13b and \opt-6.7b/13b. For instance, when considering \llama-7b, shifting from INT to FP quantization in the W8A8 configuration leads to an additional 0.25 PPL reduction (from 10.63 to 10.38), and in the W4A8 setup, there is an extra 0.4 PPL drop (from 11.48 to 11.08). These performance gains are significant, considering all other optimization parameters remain constant, and they align with the Class-3 quantization sensitivity category as defined in \cite{yao2023comprehensive}. Thus, the results underline the importance of FP8 activation, particularly in larger LLMs, to enhance the overall performance and precision of the model's outputs.

\textbf{FP8 weights rival INT8, while FP4 weights potentially outperform INT4.} From Table \ref{tab:main-fp}, we observe comparable performances between INT8 and FP8 weight quantization across various models and datasets, when keeping activation at FP8. This probably due to we used FGQ on weight quantization. Interestingly, when weight quantization is lowered, FP4 exhibits certain advantages over INT4, particularly evident in \llama-7b  (15.14 to 16.09) and \llama-13b models (11.08 to 11.31). Specifically, under the W4A8 configuration for \llama-7b, we see 0.95 improvement of FP4 over INT4, a significant gain. The preferable performance of FP4 over INT4 is particularly advantageous for hardware designs like H100, where FP8 is already supported. Thus, a simple modification to accommodate FP4 would be easier than implementing a system supporting INT4 weight and FP8 activation.






\textbf{LoRC improves W4A8.}
Table \ref{tab:main-fp} shows that the Low Rank Compensation (LoRC) method  enhanced the W4A8 quantization scheme, reducing quantization errors. This improvement is particularly pronounced in smaller models, underlining the effectiveness of LoRC in optimizing the performance of these computing processes while impacting little on the model-size.

\begin{table}[H] 
\caption{Scale values ($S$) are evaluated both without and with restrictions of being a power of 2, as shown in the second column. The quantization type employed is FP4 for weight and FP8 for activation.}\centering
\label{tab:main-scale}
\begin{adjustbox}{width=0.999\linewidth}
\centering
\begin{tabular}{ll|cc|cc|cc|cccccc }
\toprule
\multirow{2}{*}{Q-type}       &  Scale  & \multicolumn{2}{c|}{\llama-3b}  & \multicolumn{2}{c|}{\llama-7b}  & \multicolumn{2}{c|}{\llama-13b}  & \multicolumn{2}{c}{\llama-30b} \\
    &  $S=2^{n}$ & Mean & WIKI/PTB/C4  & Mean & WIKI/PTB/C4  &  Mean & WIKI/PTB/C4 &  Mean & WIKI/PTB/C4\\
\midrule
\multirow{3}{*}{W4A8}           & \xmark  & 12.45 & 7.62/20.05/9.67 & 15.14 & 6.32/31.61/7.51 & 11.08 & 5.26/21.27/6.73 & 5.92 & 4.26/7.42/6.09 \\
                             & \cmark (M1) & 12.66 & 7.76/20.41/9.81 & 16.33 & 6.34/34.82/7.82 & 10.90 & 5.31/20.63/6.78 & 6.00  & 4.38/7.48/6.15  \\
                               & \cmark (M2) & 12.55 & 7.68/20.21/9.77 & 14.49 & 6.37/29.32/7.79 & 10.95 & 5.26/20.81/6.77 &    5.97&	4.31/7.48/6.13   \\      
                             \midrule
\multirow{2}{*}{W4A8} & \xmark & 12.42 & 7.61/19.98/9.66 & 13.95 & 5.87/28.75/7.24 & 10.80 & 5.24/20.46/6.72 & 5.91 & 4.26/7.40/6.07 \\
 \multirow{2}{*}{LoRC}                             & \cmark (M1) &12.61 & 7.69/20.37/9.76 & 14.23 & 5.94/29.47/7.30 & 10.74 & 5.28/20.19/6.75 & 5.98  & 4.33/7.48/6.13 \\
 & \cmark (M2) & 12.42 & 7.63/19.89/9.74 & 13.68 & 5.90/27.83/7.32 & 10.40 & 5.23/19.22/6.76 &    5.94	&4.28/7.44/6.11  \\
\bottomrule
\end{tabular}
\end{adjustbox}
\begin{adjustbox}{width=0.999\linewidth}
\centering
\begin{tabular}{ll|cc|cc|cc|cccccc }
\toprule
\multirow{2}{*}{Q-type}       &  Scale  & \multicolumn{2}{c|}{\opt-1.3b}  & \multicolumn{2}{c|}{\opt-6.7b}  & \multicolumn{2}{c|}{\opt-13b}  & \multicolumn{2}{c}{\opt-30b} \\
    &  $S=2^{n}$ & Mean & WIKI/PTB/C4  & Mean & WIKI/PTB/C4  &  Mean & WIKI/PTB/C4 &  Mean & WIKI/PTB/C4\\
\midrule
\multirow{2}{*}{W4A8}            & \xmark & 16.29 & 15.32/18.19/15.35 & 12.09 & 10.89/13.44/11.95 & 11.34 & 10.16/12.55/11.30 & 10.72 & 9.52/11.90/10.75 \\
                             & \cmark (M1) &16.66 & 15.65/18.66/15.65 & 12.29 & 11.12/13.69/12.05 & 11.36 & 10.22/12.54/11.32 & 10.77 & 9.58/11.96/10.76 \\
                             & \cmark (M2)  & 16.47 & 15.23/18.55/15.62 & 12.25 & 11.11/13.61/12.03 & 11.40 & 10.22/12.61/11.36 & 10.74	&9.47/11.96/10.78        \\
                             \midrule
\multirow{2}{*}{W4A8}  & \xmark &16.23 & 15.50/17.92/15.28 & 12.09 & 10.96/13.40/11.90 & 11.33 & 10.15/12.55/11.29 & 10.71 & 9.48/11.90/10.75\\
 \multirow{2}{*}{LoRC}                              & \cmark (M1) & 16.47 & 15.59/18.37/15.45 & 12.17 & 11.10/13.47/11.95 & 11.36 & 10.21/12.54/11.32 & 10.74 & 9.49/11.96/10.76 \\
 & \cmark (M2) &  16.30 & 15.39/18.10/15.42 & 12.19 & 11.11/13.49/11.97 & 11.41 & 10.34/12.54/11.34 & 10.75&9.49/11.96/10.78  \\
\bottomrule
\end{tabular}
\end{adjustbox}
\end{table}

\textbf{Casting the FP4 to FP8.} As detailed in Section \ref{sec:fp4tofo8}, to maximize real latency speedup on NVIDIA H100 hardware, we suggest the scale factor $S$ for weight quantization to be represented as a power of $2$. In pursuit of this, we executed a series of experiments using FP4 for weight and FP8 for activation quantization. The results of these experiments, conducted both with and without LoRC, are presented in Table \ref{tab:main-scale}. Our data shows that while constraining the scaling factors occasionally results in unexpected improvements in models like \llama-7b and \llama-13b, we generally observe a minor degradation of quality in the W4A8 floating-point model, regardless of whether we used method M1 or M2. M2 generally outperforms M1. When we implement LoRC, this decline in quality can be mitigated, particularly in the \opt-1.3b, \llama-7b, and \llama-13b models. Hence, our results advocate for the use of LoRC, especially when considering scale restrictions for weight quantization in deep learning models.

\section{Conclusions}
\label{sec:conclusions}

In this study, we demonstrate that floating-point (FP) quantization significantly surpasses integer (INT) quantization in the context of large language models (LLMs) during post-training quantization. Notably, FP8 activation exceeds INT8, especially in larger models. Moreover, FP8 and FP4 weight quantization are either competitive with or surpass their INT equivalents. The Low Rank Compensation (LoRC) approach greatly enhances the W4A8 quantization scheme, particularly in smaller models. In conclusion, our work underscores the potential of FP quantization in enhancing model performance, and strategies such as LoRC further mitigate degradation induced by scale factor restrictions on weight quantization.

\section*{Acknowledgement}
This research was conducted within the supportive environment of the DeepSpeed team at Microsoft, whose invaluable assistance was instrumental to this project. We thank Cheng Li and Connor Homes for the insightful discussions.
{
 \bibliographystyle{plain}
\bibliography{ref.bib}
}

\appendix

\counterwithin{figure}{section}
\counterwithin{table}{section}

\section{Experiment Details}
\label{sec:experiment-details}
 As we used GPTQ method~\cite{frantar2022gptq},  we use C4 dataset to randomly select 128 sentences for the light-weight PTQ and each of them has 2048 tokens. We run them on a single GPU (i.e, V100-32GB) thanks for the two open-source github repositories.\footnote{\url{https://github.com/IST-DASLab/gptq} and \url{https://github.com/qwopqwop200/GPTQ-for-LLaMa.git}} To accommodate the real computation efficiency, the group-size for weight quantization is 256 for both model family (\opt and \llama) except the \llama-3b with 320 as its hidden-dimension is 3200. All the checkpoints we used are from huggingface.\footnote{\llama-3b is \url{openlm-research/open_llama_3b} and all other \llama are from \url{decapoda-research/llama-\#b-hf} where \url{\#} can be 7b, 13b and 30b. As for \opt, they are from \url{facebook/opt-\#b} where \url{\#} can be 1.3b, 7b, 13b and 30b. } As for activation, we perform token-wise quantization in order to accommodate the latency requirements.  For LoRC method, the dimension for the two low-rank matrix we used for \llama is 8. While for OPT, the dimension is 16, 32, 40 and 56 respectively for 1.3b, 6.7b, 13b and 30b. We did not try others dimension as indicated by \cite{yao2023comprehensive} that dimension of the low-rank matrix does not play too much impact on the quantization error as long as it larger than 8.

\begin{table}[H]
\begin{center}
\caption {Comparisons between E2M1 and E3M0. The quantization is FP4 for weight and FP8 for activation.}\label{tab:fp4}
 \begin{adjustbox}{width=0.599\linewidth}
\begin{tabular}{lcccccccccc }
\toprule
Activation (FP8) & OPT-1.3b & OPT-6.7b & OPT-13b & OPT-30b \\\midrule
Weight-FP4  (E3M0)    & 16.96    & 12.41    & 11.53   & 10.86   \\
Weight-FP4  (E2M1)    & 16.23    & 12.09    & 11.33   & 10.71   \\
\bottomrule
\end{tabular}

\end{adjustbox}
\end{center}
\end{table}

\end{document}